\documentclass{article}
\usepackage{ijcai24}

\usepackage{times}
\usepackage{soul}
\usepackage{url}
\usepackage[hidelinks]{hyperref}
\usepackage[utf8]{inputenc}
\usepackage[small]{caption}
\usepackage{graphicx}
\usepackage{amsmath}
\usepackage{amsthm}
\usepackage{booktabs}
\usepackage{algorithm}
\usepackage[switch]{lineno}

\usepackage{lineno,hyperref}
\usepackage{graphicx}
\usepackage{epstopdf}
\usepackage{algorithm}
\usepackage{algorithmicx}
\usepackage{algpseudocode}
\usepackage{amsmath}
\usepackage{amstext}
\usepackage{amssymb}
\usepackage{caption}
\usepackage{subfigure}
\usepackage{color}
\usepackage{multirow}
\usepackage{appendix}
\usepackage{stfloats}

\title{Blockchain-enabled Trustworthy Federated Unlearning}

\author{
Yijing Lin$^{1,3}$\and
Zhipeng Gao$^{1}$\thanks{Corresponding author: Zhipeng Gao.}\and
Hongyang Du$^{2}$\and
Jinke Ren$^{3,4}$\and
Zhiqiang Xie$^{1}$\and
Dusit Niyato$^{2}$ 
\affiliations
\vspace{0.5em}
\normalsize{$^1$ State Key Laboratory of Networking and Switching Technology, Beijing University of Posts and Telecommunications\\
$^2$ School of Computer Science and Engineering, Nanyang Technological University\\
$^3$ The Future Network of Intelligence Institute, The Chinese University of Hong Kong (Shenzhen)\\
$^4$ School of Science and Engineering, The Chinese University of Hong Kong (Shenzhen)\\}
\emails \tt{\normalsize{gaozhipeng@bupt.edu.cn}}
}

\begin{document}

\maketitle

\begin{abstract}

Federated unlearning is a promising paradigm for protecting the data ownership of distributed clients. It allows central servers to remove historical data effects within the machine learning model as well as address the ``right to be forgotten" issue in federated learning. However, existing works require central servers to retain the historical model parameters from distributed clients, such that allows the central server to utilize these parameters for further training even, after the clients exit the training process. To address this issue, this paper proposes a new blockchain-enabled trustworthy federated unlearning framework. We first design a proof of federated unlearning protocol, which utilizes the Chameleon hash function to verify data removal and eliminate the data contributions stored in other clients' models. Then, an adaptive contribution-based retraining mechanism is developed to reduce the computational overhead and significantly improve the training efficiency. Extensive experiments demonstrate that the proposed framework can achieve a better data removal effect than the state-of-the-art frameworks, marking a significant stride towards trustworthy federated unlearning.

\end{abstract}

\section{Introduction}

The astonishing success of AI-generated content (AIGC) has led to a resurgence in the popularity of machine learning (ML) technologies. However, the performance of ML models relies heavily on a large volume of data from massive distributed clients. On the other hand, many international regulations such as the General Data Protection Regulation (GDPR)~\cite{voigt2017eu} have stipulated that ML service providers are obligated to ensure the ``right to be forgotten" for clients, i.e., allowing them to remove their data effects from well-trained models. This necessity gives rise to a new learning paradigm called \textit{machine unlearning}, which can remove data effect of target clients from the learning models without retraining from scratch~\cite{xu2023machine}. Specifically, given an unlearning request, the well-trained model will run a pre-defined unlearning algorithm to forget the data that is used in the training process.

Machine unlearning holds great potential in protecting data privacy, while its implementation in practice faces a key challenge of high unlearning costs. Previous works adopt a joint sharded, isolated, sliced, and aggregated (SISA) training approach~\cite{bourtoule2021machine}, which uses data sharding and slicing on clients to reduce computational overhead. However, in practical distributed learning paradigms, e.g., federated learning (FL), data is held by decentralized clients. The clients collaboratively train a global model by uploading local models to a central server for aggregation~\cite{mcmahan2017communication}. Since FL shares model updates instead of raw data, traditional ML unlearning methods requiring direct access to data cannot be applied~\cite{wang2022federated}. Moreover, the unlearning of specific model updates from the global model is quite complex, as the contribution of individual clients cannot be isolated easily. 

Existing works have utilized storage-and-calibration~\cite{wu2022federated}, Newton-type model update~\cite{liu2022right}, model pruning~\cite{wang2022federated}, and reverse stochastic gradient ascent~\cite{wu2022federated} algorithms to achieve client-level data removal in federated unlearning. However, these solutions still suffer from two shortcomings. First, there exist some irremovable parameters controlled by the central server, which may be utilized for further model training without permission, thereby compromising the data ownership of the target clients~\cite{wang2023federated}. Therefore, the verification of whether the central server employs the target clients' parameters for aggregation becomes quite challenging. Second, it is difficult to determine the optimal unlearning rounds and reduce computation overhead. To address these issues, we propose a blockchain-enabled trustworthy federated unlearning framework to verify the ``right to be forgotten" in a decentralized manner. Within the framework, we design a Chameleon hash function-based proof of federated unlearning to eliminate the data effects of the target clients. The Chameleon hash function is quite useful in our framework due to its collision resistance and key-exposure freshness properties. It also allows authorized data changes without altering the hash value, thus ensuring flexible and secure management of data without unnecessary retraining. To reduce the number of unlearning rounds and minimize the computational overhead, we further develop an adaptive retraining mechanism to evaluate the specific contributions of the target clients. In summary, the contributions of this work can be illustrated as follows:

\begin{itemize}
    \item We propose a blockchain-enabled trustworthy federated unlearning framework, which utilizes on-chain smart contracts and off-chain hash mappings to seamlessly handle continuous unlearning requests and verify data removal from the global FL models.    
    \item To ensure the ``right to be forgotten" of clients, we design a Chameleon hash function-based proof of federated unlearning protocol. In particular, the target clients can invoke an unlearning rewriting function to fully erase model updates and data effects associated with the target clients.
    \item To determine the number of unlearning rounds, we propose an adaptive contribution-based retraining mechanism. By quantifying the contributions of target clients by historical model updates and estimating the unlearning rounds, the overhead can be significantly reduced.    
    \item Experimental results show that the proposed framework can achieve better performance in terms of accuracy and security as compared with the three benchmark frameworks.
\end{itemize}

\section{Related Work}

\subsection{Machine Unleaning}

Machine unlearning is able to eliminate the data effects from ML models without requiring retraining from scratch. Specifically, when receiving a data removal request, the ML model will execute a pre-defined unlearning mechanism to erase the associated data effects involved in the model. To improve the effectiveness of machine unlearning, ~\cite{chundawat2023can} proposes a student-teacher framework, which includes a competent teacher and an incompetent teacher to selectively transfer knowledge and deliberately exclude information related to the target data. ~\cite{pan2022machine} designs a federated K-means clustering algorithm for efficient machine unlearning, and develops a sparse compressed multiset aggregation mechanism to reduce communication overhead. ~\cite{bourtoule2021machine} partitions all data samples into several distinct shards, trains separate models on each shard, and uses slicing methods to minimize the computational overhead. On the other hand, to test the unlearning effectiveness of machine unlearning,~\cite{chen2021machine} utilizes the membership inference attack to predict whether the target data belongs to the training data. While these approaches have greatly improved the efficiency and security of machine unlearning, they cannot be directly used in FL since FL requires periodical exchanging of model updates rather than raw data. Therefore, new solutions are needed to adapt machine unlearning in FL.

\subsection{Federated Unlearning}

FL is a distributed learning paradigm for protecting data privacy. Due to the characteristics of decentralized data storage and indirect data transmission inherent to FL, it requires new solutions to achieve efficient data removal. FedEraser~\cite{liu2021federaser} introduces a storage-and-calibration mechanism to eliminate the model updates of target clients in the global calibration rounds, thus making the aggregated global model forget the corresponding data effects. For evaluating the effectiveness of federated unlearning,~\cite{gao2022verifi} employs the watermark and fingerprint on the models of the target clients, and compares the performance on marked data before and after unlearning. For rapid retraining, ~\cite{liu2022right} introduces a distributed Newton-type model update algorithm to approximate the loss function utilized for calibration. By employing the diagonal empirical Fisher information matrix, it effectively reduces the computational cost associated with calculating the inverse Hessian matrix. To selectively forget categories from FL models,~\cite{wang2022federated} propose a term requency Inverse Document Frequency-based federated unlearning method, which evaluates the channel contribution for model pruning and class discrimination. To facilitate the removal of the target training data,~\cite{wu2022federated} proposes a federated unlearning framework based on the reverse stochastic gradient ascent algorithm. However, the aforementioned works neither address the challenge of removing parameters controlled by the central server nor determine the required unlearning rounds.

\subsection{Blockchain-based Proof of Learning}

Blockchain is a promising technology to protect data integrity and realize decentralization for FL. In particular, ~\cite{lan2021proof} proposes a proof of work based decentralized learning mechanism, which utilizes computational power and data encryption to train neural networks and maintain data integrity.~\cite{lin2022novel} utilizes blockchain data and computing oracle to achieve efficient data interactions and computation between blockchain and FL.~\cite{chowdhury2023sf} utilizes blockchain and off-chain storage to fetch the client's model updates and assesses the quality of the client's and the global model updates to prevent malicious clients.  Additionally, to optimize communication efficiency,~\cite{cui2022fast} develops a fast blockchain-based FL framework by dynamically adjusting the compression rate and block generation rate. Moreover,~\cite{lin2023drl} utilizes deep reinforcement learning to design an adaptive blockchain sharding mechanism to improve the efficiency of decentralized FL. Despite the great success of the above works, the potential applications of blockchain in federated unlearning still remain unexplored.

\begin{figure*}[!t]
    \centering
    \includegraphics[width=1.0\textwidth]{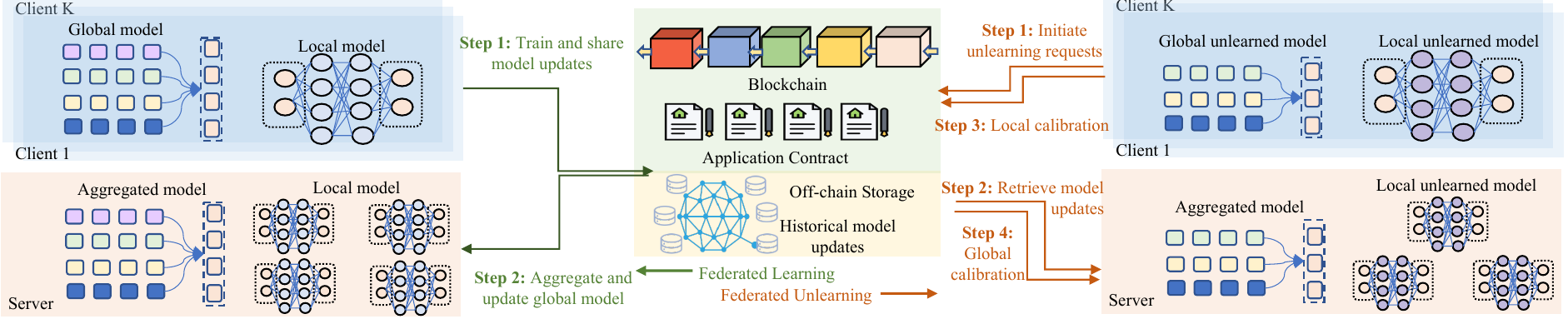}
    \caption{Blockchain-enabled trustworthy federated unlearning framework.}
    \label{fig:framework}
\end{figure*}

\section{Framework}

In contrast to traditional learning frameworks that collect data for centralized training, FL only exchanges model updates between clients and the server to preserve data privacy. Nevertheless, this unique characteristic makes it hard to remove data effects when training data cannot be accessed globally~\cite{wang2022federated}. To address this issue, we propose a blockchain-enabled trustworthy federated unlearning framework with $K$ clients and a server selected by consensus, as shown in Figure~\ref{fig:framework}. It consists of two parts, i.e., an on-chain component and an off-chain component, which execute unlearning requests and store model updates, respectively.

\textbf{On-chain component} comprises the operations that modify the network states of the blockchain, including the learning and unlearning requests, the uploaded hashes of local model from clients, and the hashes of aggregated model updates from the server. These network states are recorded on the blockchain through smart contracts, which require consensus within the network. 

\textbf{Off-chain component} includes the operations that do not alter the blockchain network states, such as training local models on the clients and aggregating models on the server. For illustration purposes, we use the green and purple lines (in Figure~\ref{fig:framework}) to show the workflows of FL and unlearning.

\paragraph{Workflow of blockchain-enabled FL:} For the training round $t$, each client $k \in \{1,\ldots, K\}$ starts by receiving the current global model ${\mathbf{M}}^t$ from the server, where the server is selected by consensus. \textbf{Step 1:} Each client $k$ uses its local dataset $\mathcal{D}_k$ to train its local model update $\mathbf{U}_k^t$. Next, each client shares its $\mathbf{U}_k^t$ with the off-chain storage and secures a hash $H_k^t$ that associates $\mathbf{U}_k^t$ with the blockchain. The hash $H_k^t$ is written into a smart contract for consensus and is broadcast to the selected server. $\textbf{Step 2:}$ The server collects all the hashes $\mathcal{H}^t=\{H_1^t,\ldots,H_K^t\}$ from the blockchain, aggregates local updates $\mathcal{U}^t=\{\mathbf{U}_1^t,\ldots,\mathbf{U}_K^t\}$ from clients, and updates the global model. Finally, the updated model $\mathbf{M}^{t+1}$ is stored in off-chain storage, while the corresponding mapping hash $H^{t+1}$ is recorded within the smart contract and broadcast to the clients afterward. These steps will repeated for $T$ rounds until achieving the optimal performance.

\paragraph{Workflow of blockchain-enabled federated unlearning:} $\textbf{Step 1:}$ In the $t_u$-th unlearning round, the target clients first issue their unlearning requests on the smart contract and send them to the selected server. The unlearning requests are used to remove the parameters trained by the clients' data from the aggregated model $\mathbf{M}^{t_u}$. $\textbf{Step 2:}$ Upon receiving the requests, the selected server collects all hash $\mathcal{H}^{t_u+1}=\{H_1^{t_u},\ldots,H_K^{t_u}\}\backslash \mathcal{H}_{k_u}^{t_u}$ except the ones corresponding to the target clients. Then the server retrieves the local updates $\mathcal{U}^{t_u}=\{\mathbf{U}_1^{t_u},\ldots,\mathbf{U}_K^{t_u}\}\backslash \mathcal{U}_{k_u}^{t_u}$ and aggregates them by

\begin{equation}
    \small
    \mathbf{U}^{t_u+1}=\frac{\sum_{k=1,k \neq k_u}^K w_k \mathbf{U}_k^{t_u}}{(K-1) \sum_{k=1,k \neq k_u}^K w_k},
    \label{eq:calibration}
\end{equation} where $w_k$ is the weight of client $k$. Also, the server updates the global model, denoted by $\mathbf{M}^{t_u+1}$, and stores it on the off-chain storage. Subsequently, the hash $H^{t_u+1}$ is shared with the clients that remain in the process. $\textbf{Step 3:}$ The retained clients compute their local model updates $\mathbf{U}_k^{t_u+1}$ for calibration. $\textbf{Step 4:}$ The server collects local model updates from the retained clients and performs global calibration by

\begin{equation}
    \small
    \mathbf{M}^{t_u+1}=\mathbf{M}^{t_u}+\mathbf{U}^{t_u+1}.
\end{equation}

Due to the traceability of blockchain, the target clients can obtain the global model updates $\mathbf{U}^{t_u+1}$ for calibration, thus allowing the verification of whether $\mathbf{U}^{t_u+1}$ incorporates their model updates. Specifically, the target clients query the calibration hash $H^{t_u+1}$ and the hashes from the retained clients, and retrieve the model updates $\mathcal{U}^{t_u}$. The target clients can calculate the output model updates according to (\ref{eq:calibration}) and determine whether the output matches the expected value $U^{t_u+1}$ to achieve unlearning verification.

\section{Proof of Federated Unlearning Protocol} 

In the above two workflows, the shared model updates of the target clients can be queried by other clients and the servers, even after the target clients have exited the training process. As a result, those model updates may be used in some unauthorized activities, such as adding to the retraining process. To mitigate this risk, we propose a Chameleon hash function-based proof of federated unlearning protocol to protect the data ownership of the target clients. 

\subsection{Protocol Implementation}

The protocol consists of five phases, including initialization, key generation, parameter sharing, unlearning verification, and unlearning rewriting.

\paragraph{Initialization:} First, the clients generate system configurations of the Chameleon hash for unlearning. Specifically, the clients respectively generate two large prime numbers $p$ and $q$. To ensure that $q$ divides $p-1$, we have 
\begin{equation}
    \small
    p = nq + 1, n \in \mathbb{N}^*.
\end{equation} Then the clients determine the multiplicative group of integers modulo $p$, denoted by $Z_p^*$. We note that the order of this group is given by $q$, signifying the number of elements in the group. Finally, the clients identify a generator $g$ for the group\footnote{A generator of a group is a specific element that can be utilized to generate all other elements of the group through its powers.}. In the group, each non-zero element can be generated by $g^x$, where $x$ is a random integer.

\paragraph{Key Generation:} Given a security parameter $\lambda$ and a random integer $x$, the clients generate a public key $\mathsf{pk}$ and a private key $\mathsf{sk}$ by
\begin{equation}
    \small
    \mathsf{pk}=(g,h),\quad \mathsf{sk}=x,
\end{equation} where $h=g^x \mod p$. For brevity, we define the key generation function $\mathsf{ChGen}(1^{\lambda})=(\mathsf{pk},\mathsf{sk})$. Then, instead of using a conventional hash function, the clients send the public key to the blockchain. This allows other clients to compute the Chameleon hash, which is essential for performing federated unlearning on the shared model updates. 

\paragraph{Parameter Sharing:} Each client $k$ store local model updates $\mathbf{U}_k^t$ on the off-chain storage and generate a corresponding Chameleon hash $H_k^t$ in the training round $t$. The hash generation function is defined as $\mathsf{ChHash}(\mathsf{pk},\mathbf{U}_k^t,r)=g^{\mathbf{U}_k^t} h^r \mod p$. The random value $r$ is a cryptographic blinding factor that adds an additional layer of randomness to enhance the security of the Chameleon hash. For example, when the secret key $\mathsf{sk}$ is unavailable, it is computationally infeasible to find two different model updates $\mathbf{U}_{k_1}^t$, $\mathbf{U}_{k_2}^t$ and random values $r_1$, $r_2$ such that $\mathsf{ChHash}(\mathsf{pk},\mathbf{U}_{k_1}^t,r_1)=\mathsf{ChHash}(\mathsf{pk},\mathbf{U}_{k_2}^t,r_2)$. Thereafter, the Chameleon hash $H_k^t$ is recorded into the smart contracts to confirm the participation of client $k$.

\paragraph{Unlearning Verification:} When target clients send an unlearning request to the blockchain, the selected server first collects the relevant Chameleon hashes to access the model updates of the retained clients. Then, it performs the calibration process i.e., (\ref{eq:calibration}), to generate the calibration hashes, as 

\begin{equation}
    \small
    H^{t_u+1}=\mathsf{ChHash}(\mathsf{pk},\mathbf{U}^{t_u+1},r)=g^{\mathbf{U}^{t_u+1}} h^r \mod p.
\end{equation} Next, the target clients verify the correctness of $H^{t_u+1}$ by calling the verification function $\mathsf{ChVerify}(\mathsf{pk},\mathbf{U}^{t_u+1},H^{t_u+1},r)=1/0$. If the output is one, the process is deemed correct; otherwise, it is wrong. The target clients then use the hashes stored on the blockchain to check whether the retraining process has utilized its model updates for model calibration and data removal. Specifically, they retrieve other model updates through the stored hashes $\mathcal{H}^{t_u+1}=\{H_1^{t_u},\ldots,H_K^{t_u}\}\backslash \mathcal{H}_{k_u}^{t_u}$ and computes the aggregated model updates $\hat{\mathbf{U}}^{t_u+1}$ using (\ref{eq:calibration}). After that, they generates the hash $\hat{H}^{t_u+1}=\mathsf{ChHash}(\mathsf{pk},\hat{\mathbf{U}}^{t_u+1},r)=g^{\hat{\mathbf{U}}^{t_u+1}} h^r \mod p$ and verify whether $H^{t_u+1}$ equals $\hat{H}^{t_u+1}$. Since the on-chain hashes and off-chain model updates are assured by the blockchain and are accessible by other clients and servers, this unlearning verification process can be independently verified by other clients as well.

\paragraph{Unlearning Rewriting:} As mentioned earlier, the model updates of the target clients are always accessible to other clients and may be used in the subsequent training process. This poses a severe privacy risk, necessitating the removal of the model updates while preserving the historical participation records. To address this issue, each target client first generates random model updates, denoted as $\overline{\mathbf{U}}_{k_u}^{t_u}$, to replace the real model updates with the same hash, ensuring that the Chameleon hash verifications for both the original and random updates yield the same result. In this way, the client's data contribution can be erased without disrupting the on-chain hash of the whole system. The above process can be expressed as 

\begin{equation}
    \small
    \left\{
    \begin{split}
        &\mathsf{ChRewrite}(\mathsf{sk}, \mathbf{U}_{k_u}^{t_u}, \overline{\mathbf{U}}_{k_u}^{t_u},H_{k_u}^{t_u})=1, \\
        &\mathsf{ChVerify}(\mathsf{pk},\overline{\mathbf{U}}_{k_u}^{t_u},H_{k_u}^{t_u},\overline{r})=\mathsf{ChVerify}(\mathsf{pk},\mathbf{U}_{k_u}^{t_u},H_{k_u}^{t_u},r), \\
        & \mathsf{ChHash}(\mathsf{pk},\mathbf{U}_{k_u}^{t_u},r)=\mathsf{ChHash}(\mathsf{pk},\overline{\mathbf{U}}_{k_u}^{t_u},\overline{r}),
    \end{split}
    \right.
\end{equation} where $\overline{r}=\frac{\mathbf{U}_{k_u}^{t_u}-\overline{\mathbf{U}}_{k_u}^{t_u}}{x}+r \mod p$ ensures that the off-chain model updates are replaced by random values, completing the unlearning process of the client's data. The output of $\mathsf{ChRewrite}$ represents whether the rewriting operations are successful.

\subsection{Security Analysis} 

Without loss of generality, we use two widely employed metrics to evaluate the security of the proposed proof of federated unlearning protocol, including the collision resistance and the key exposure freshness.

\paragraph{Collision Resistance} refers to the property that it is computationally infeasible for an adversary $\mathsf{A}$ to find two different inputs that hash to the same value when the secret key $\mathsf{sk}$ is unavailable. This property ensures that each hash mapping model update can serve as a unique fingerprint that cannot be replaced without the correct secret keys. Mathematically, the collision resistance can be expressed as

\begin{equation}
    \small
    \Pr \left[
    \begin{array}{c}
      \mathsf{ChGen}(1^{\lambda})=(\mathsf{pk},\mathsf{sk}) \\
      \mathsf{ChRewrite}(\mathsf{sk}_{\mathsf{A}}, \mathbf{U}_{k_u}^{t_u}, \overline{\mathbf{U}}_{k_\mathsf{A}}^{t_u},H_{k_u}^{t_u})=1 \\ 
      \mathsf{ChRewrite}(\mathsf{sk}, \mathbf{U}_{k_u}^{t_u}, \overline{\mathbf{U}}_{k_u}^{t_u},H_{k_u}^{t_u})=1 \\
      \mathsf{ChVerify}(\mathsf{pk},\mathbf{U}_{k_u}^{t_u},H_{k_u}^{t_u},r)=1 \\
      \mathsf{ChVerify}(\mathsf{pk},\mathbf{U}_{k_\mathsf{A}}^{t_u},H_{k_u}^{t_u},r)=1
    \end{array}
\right] \leq \mathsf{negl}(\lambda_1),
    \label{eq:collision}
\end{equation} where the second and fifth equations of (\ref{eq:collision}) represent the unauthorized rewrite attempt by the adversary and the verification of the adversary's input, respectively. $\mathsf{negl}(\lambda_1)$ means that it is negligible for the adversary to attack the protocol successfully with respect to the security parameter $\lambda_1$.

\paragraph{Key-exposure Freshness} means that the exposure of the secret key $\mathsf{sk}$ does not compromise the security of the previous hash values mapping model updates. Even if an adversary $\mathsf{A}$ obtains the secret key of the target client, it cannot retroactively tamper or forge past hash values that have been recorded on the blockchain, hence protecting the historical model updates. Mathematically, the key-exposure freshness can be expressed as

\begin{equation}
    \small
    \Pr \left[
    \begin{array}{c}
    \mathsf{ChGen}(1^{\lambda})=(\mathsf{pk},\mathsf{sk}) \\
    \mathsf{ChHash}(\mathsf{pk},\mathbf{U}_{k_u}^{t_u},r)=H_{k_u}^{t_u} \\
    \mathsf{ChVerify}(\mathsf{pk},\mathbf{U}_{k_u}^{t_u},H_{k_u}^{t_u},r)=1 \\
    \mathsf{ChRewrite}(\mathsf{sk}, \mathbf{U}_{k_u}^{t_u}, \overline{\mathbf{U}}_{k_\mathsf{A}}^{t_u},H_{k_u}^{t_u}) \\
    \mathsf{ChVerify}(\mathsf{pk},\overline{\mathbf{U}}_{k_\mathsf{A}}^{t_u},H_{k_u}^{t_u},\overline{r})\neq 1
    \end{array}
    \right] \leq \mathsf{negl}(\lambda_2),
    \label{eq:key-exposure}
\end{equation} where the third and fourth equations of (\ref{eq:key-exposure}) represent the successful verification with the correct model updates and the failed unauthorized rewrite attempt by the malicious client or server. Similarly, $\mathsf{negl}(\lambda_2)$ reflects that it is negligible for the adversary to achieve successful unauthorized rewriting with respect to the security parameter $\lambda_2$.

\section{Adaptive Retraining Mechanism}

While the proof of federated unlearning protocol enables unlearning verification, the retained clients still need to recalibrate the models to remove the data effects of the target clients. However, it is challenging since the optimal number of unlearning rounds is hard to determine. To solve this problem, we further introduce an adaptive retraining mechanism to achieve flexible retraining of calibration rounds without compromising accuracy.

\subsection{Unlearning Retraining}
\label{sec:adaptive_retrain}

In our framework, the servers cannot access the raw data, but the contributions of the target clients can be determined by the model updates sharing with blockchain~\cite{wu2021fast}. Specifically, the convergence upper bound of FL after $T$ global rounds is given by:

\begin{equation}
    \small
    \begin{split}
        & F(\mathbf{w}(t+1)) \leq F(\mathbf{w}(t))\\
        & -\eta \sum_{t=0}^{T-1} \mathbb{E}_{k|t} \Bigg[\left(\frac{\langle \nabla F(\mathbf{w}(t)), \nabla F_k(\mathbf{w}(t)) \rangle}{\Vert \nabla F(\mathbf{w}(t)) \Vert \Vert \nabla F_k(\mathbf{w}(t)) \Vert}-\frac{B\beta\eta}{2} \right) \\
        & \times \frac{A^2}{B} \Vert \nabla F(\mathbf{w}(t)) \Vert^2 \Bigg],
    \end{split}
    \label{eq:converge}
\end{equation} where $\mathbf{w}$ is the global model, $F(\mathbf{w})$ is the global objective function, $F_k(\mathbf{w})$ is the local objective function of client $k$, $\nabla$ is the gradient operator, $\langle \rangle$ is the inner product operator, and $\Vert \Vert$ is the L2 norm. Each model update can be computed by: $\mathbf{U}_k^t=\mathbf{w}_k(t)-\mathbf{w}(t-1)$, $\eta$, $\beta$, $A$, and $B$ are denoted the learning rate, Lipschitz constant, and two parameters that associated with the bounds on local dissimilarity for clients. The expectation operation $\mathbb{E}_{k|t}$ is taken over client $k$ in round $t$. 

We can observe from (\ref{eq:converge}) that the correlation between the local gradient $\nabla F_k(\mathbf{w}(t))$ and the global gradient $\nabla F(\mathbf{w}(t))$ can serve as an effective metric for evaluating the contributions of the target client $k$ in the round $t$, which can be quantified by 

\begin{equation}
    \small
    \theta_k^t = \arccos \frac{\langle \nabla F(\mathbf{w}(t)), \nabla F_k(\mathbf{w}(t)) \rangle}{\Vert \nabla F(\mathbf{w}(t)) \Vert \Vert \nabla F_k(\mathbf{w}(t)) \Vert}.
\end{equation} 

Due to the instability of each training round, it is hard to compute the instantaneous $\theta_k^t$. As an alternative, we use the historical quantities and approximate $\theta_k^t$ by

\begin{equation}
    \small
    \begin{split}
        \widetilde{\theta}_k^t = \left\{
            \begin{aligned}
            &\theta_k^t, & t=1, \\
            &\frac{t-1}{t} \widetilde{\theta}_k^{t-1} + \frac{1}{t}\theta_k^t, & t>1. \\
            \end{aligned}
            \right.
    \end{split}
    \label{eq:correlation}
\end{equation}

We employ the Gompertz function~\cite{gibbs2000variational} to accommodate both the initial rapid changes and the eventual saturation of the contribution from the target clients. The contribution of each target client $k$ is given by 

\begin{equation}
    \small
    f(\widetilde{\theta}_k^t)=\alpha(1-e^{-\alpha e^{ (\widetilde{\theta}_k^t-1)})},
\end{equation} where $\alpha$ is a constant that controls the decreasing rate of $f(\widetilde{\theta}_k^t)$. Based on this, the contributions of the target client $k$ can be discerned by mapping the correlation between the local and global model updates. Therefore, the unlearning round is given by

\begin{equation}
    \small
    \begin{split}
        \widetilde{T}=(1-\frac{f(\widetilde{\theta}_k^t)}{\sum_{k=1,k \neq k_u}^K f(\widetilde{\theta}_k^t)})T.
    \end{split}
    \label{eq:time}
\end{equation}

In summary, we describe the proposed federated unlearning process in Algorithm~\ref{algo:workflow}.

\begin{algorithm}[!t]
  \caption{{\small Blockchain-enabled Trustworthy Federated Unlearning}}
  \small
  \begin{algorithmic}[1]
    \State Key Generation and Parameter Sharing
    \State \textbf{Unlearning Verification:}
    \State Each client sends its federated unlearning request to blockchain.
    \State The server collects relevant hashes $\mathcal{H}^{t_u+1}$.
    \State The server performs global calibration according to (\ref{eq:calibration}) and generate $H^{t_u+1}$. 
    \State The target clients verify the correctness of $H^{t_u+1}$.
    \State \textbf{Unlearning Rewriting:}
    \State The target clients replace the model updates and determine the unlearning rounds $\widetilde{T}$ according to (\ref{eq:time}).
    \For{each global unlearning round $t \in \{1,\ldots,\widetilde{T}\}$}
        \State The server perform the global calibration according to (\ref{eq:calibration}).
        \State The retained clients perform local calibration.
        \State The target clients verify the correctness.
    \EndFor
  \end{algorithmic}  
  \label{algo:workflow}
\end{algorithm}

\subsection{Time Analysis}
\label{sec:time_analysis}

According to Section~\ref{sec:adaptive_retrain}, the proposed contribution-based adaptive retraining mechanism can reduce the global rounds by $\frac{f(\widetilde{\theta}_k^t)}{\sum_{k=1,k \neq k_u}^K f(\widetilde{\theta}_k^t)}T$. To further improve the computational efficiency of federated unlearning, we introduce two hyperparameters $\delta_t$ and $c$, where the former is used to adjust the size of the retraining model update and the latter is a calibration ratio for reducing the number of local training epochs. This strategy can reduce the overall time by a factor of $\dfrac{\delta_t}{c}$ compared to the vanilla strategy of retraining from scratch. Since the unlearning rewriting operation of the target clients and the retraining operation of the retained clients are independent, the total time reduction of the proposed mechanism is given by $\dfrac{\delta_t}{c}(T-\widetilde{T})$, yielding a substantial efficiency gain.

\section{Experiments}

\subsection{Experimental Setting}

\paragraph{Blockchain.} The on-chain component is deployed using the Xuperchain v3.10.3\footnote{https://github.com/xuperchain/xuperchain} and xuper-sdk-go\footnote{github.com/xuperchain/xuper-sdk-go/v2/} developed by Baidu. The smart contract is written in Golang 1.20.2 and the default contract execution time is set as 500 ms. The port 37101 is designated as the TCP server to interact with the off-chain component, which provides an efficient interface for communication. For consensus, we use two popular algorithms: the delegated proof of stake (DPoS)~\cite{larimer2014delegated} and the proof of work (PoW)~\cite{nakamoto2008bitcoin}. DPoS improves scalability by allowing stakeholders to vote on a selected number of block validators. PoW secures the network by requiring miners to perform complex computations, thus validating and recording transactions on the blockchain. It should be noted that our framework is compatible with various consensus algorithms. Moreover, we set the initial mining difficulty level of the PoW algorithm as 19, with adjustments to the difficulty made every 10 blocks to maintain a consistent block creation rate. 

\paragraph{Unlearning.} We implement the off-chain component on a computer with Ubuntu 16.04.7 LTS, an Intel(R) Xeon(R) CPU E5-2620v4 @ 2.10GHz with 8 cores, 64GB memory, and an NVIDIA RTX 3090 GPU. The unlearning process is performed by Pytorch 1.11.0 and Torchvision 0.12.0. Moreover, we use three real datasets for the experiment, i.e., MNIST~\cite{lecun1998gradient}, Fashion-MNIST (FMNIST)~\cite{xiao2017fashion}, and CIFAR-10 (CIFAR)~\cite{krizhevsky2009learning}. The learning model associated with MNIST and FMNIST is a CNN with two convolutional layers (20 and 50 channels, 5$\times$5 kernels) and two max-pooling layers (2$\times$2 windows), followed by two fully connected layers (500 and 10 neurons). Besides, the learning model associated with CIFAR is a CNN with two convolutional layers (6 and 16 channels, 5$\times$5 kernels) each followed by a max-pooling layer (2$\times$2 windows), and three fully-connected layers (120, 84, and 10 neurons), interleaved with a dropout layer. The experimental setup for unlearning includes 50 clients, a calibration ratio of 0.5, a time interval of 2, 40 global rounds, 10 local epochs, and a local learning rate of 0.1.

\paragraph{Baselines.} For comparison purposes, we consider three baseline frameworks: 1) FedAvg~\cite{mcmahan2017communication}, a vanilla FL framework that shares model updates to preserve data privacy; 2) FedEraser~\cite{liu2021federaser}, a pioneering framework that achieves federated unlearning by collecting model updates and calibrating the aggregated model; 3) RapidTrain~\cite{liu2022right}, the state-of-the-art framework that employs a distributed Newton-type model update algorithm for fast retraining. 

\paragraph{Evaluation Metrics.} We use four performance metrics: accuracy, loss, precision, and recall of membership inference attacks (MIAs)~\cite{shokri2017membership}. Accuracy and loss are used to evaluate the inference performance of the unlearned models. The attack precision in MIAs measures the percentage of correctly identified members among those inferred as members. The recall of MIAs calculates the percentage of actual members that are correctly identified.

\subsection{Performance Evaluation}

\begin{figure}[!t]
    \centering
    \subfigure[DPoS consensus]{\includegraphics[width=1.65in]{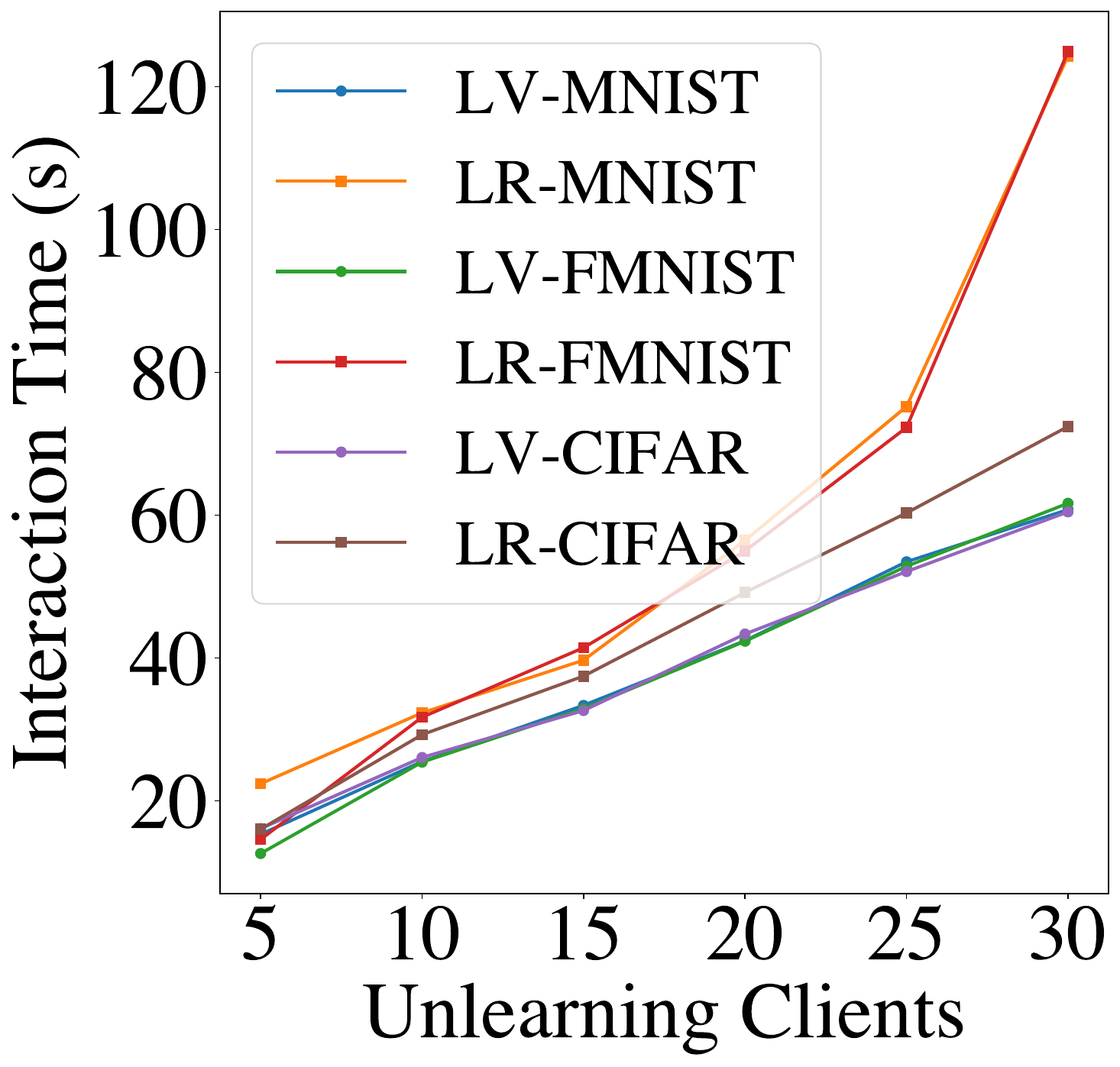}}
    \subfigure[PoW consensus]{\includegraphics[width=1.65in]{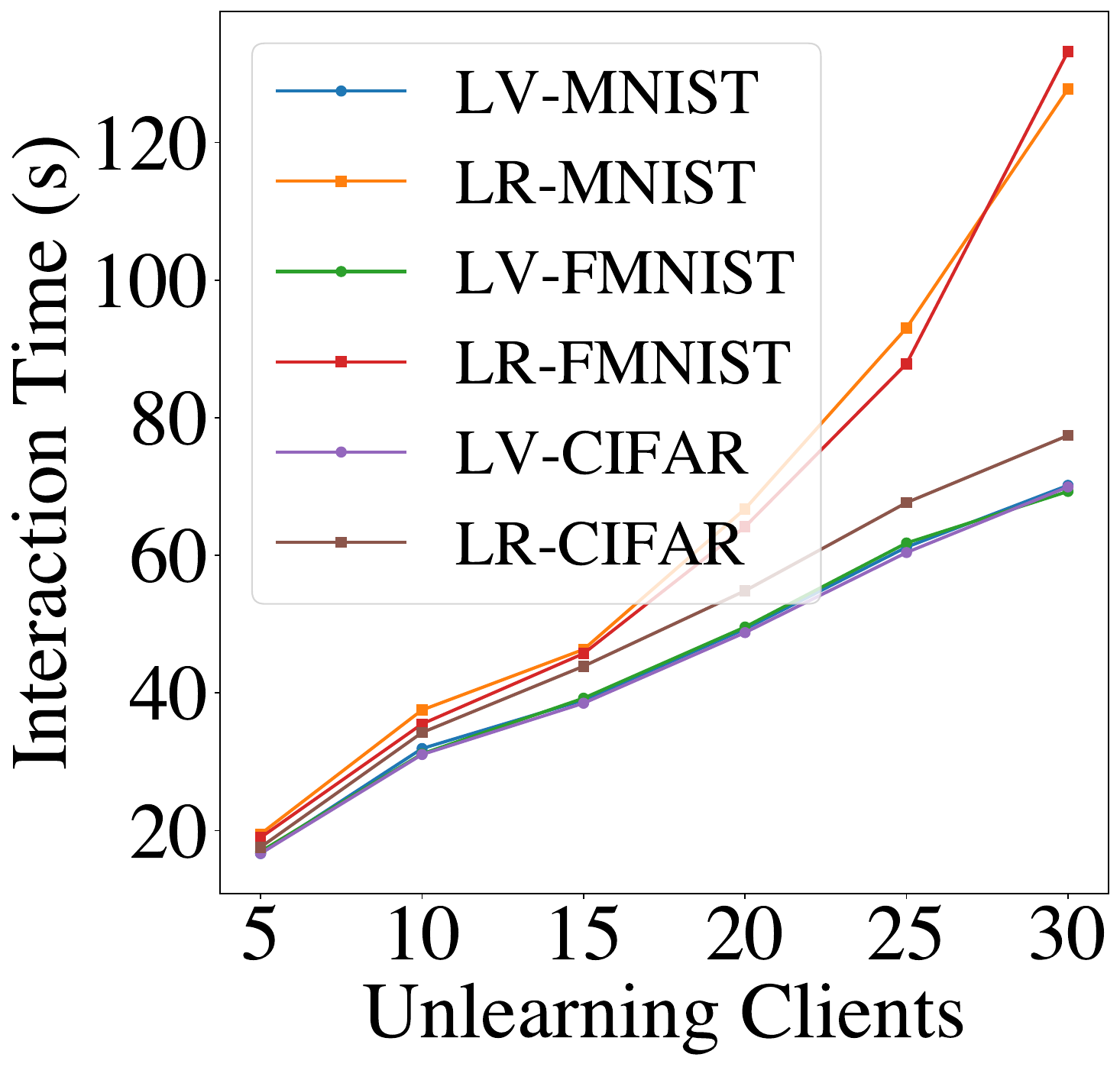}}
    \caption{On-chain and off-chain interaction time.}
    \label{fig:time}
\end{figure}

\paragraph{On-chain and Off-chain:} Figure~\ref{fig:time} illustrates the interaction times for both on-chain and off-chain components under the DPoS and PoW consensus algorithms. In particular, $\mathsf{LV}$ and $\mathsf{LR}$ correspond to the operations of local model update verification and replacement, respectively, in the context of unlearning verification and rewrite processes. According to Figure~\ref{fig:time}, we can see that the interaction time of operations increases with the number of unlearning clients. Interestingly, for different consensus algorithms, the interaction time remains almost unchanged. Therefore, we can conclude that the computation times for hash verification and replacement dominate the interaction time. Since the unlearning clients do not participate in the calibration process, the hash verification and replacement operations are independent of the retraining process. Therefore, the proposed proof of federated unlearning protocol can preserve privacy without affecting the retraining process.

\begin{figure}[!t]
  \centering
  \includegraphics[width=3.5in]{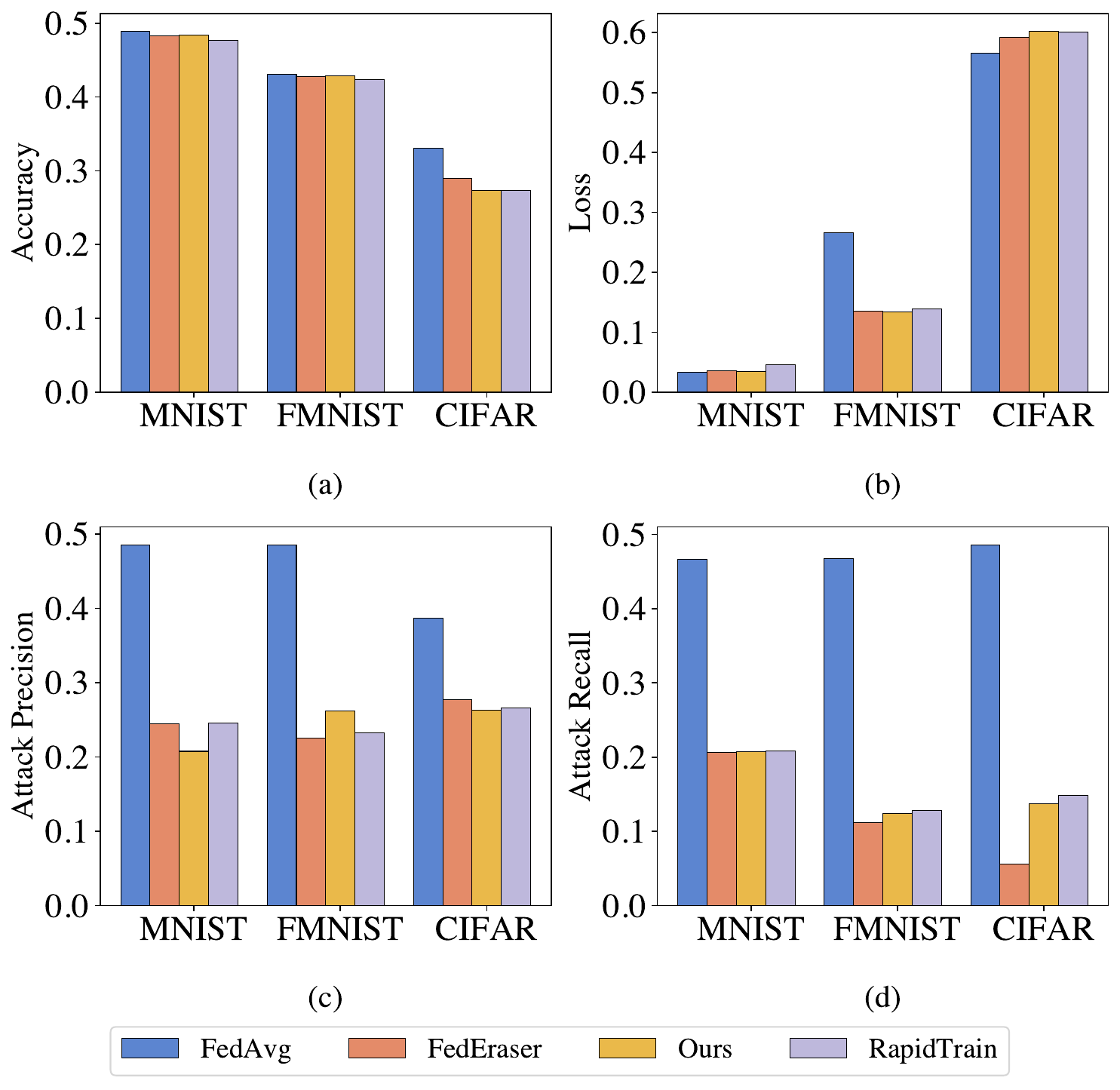}
  \caption{Accuracy and unlearning performance.}
  \label{fig:accuracy}
\end{figure}

\paragraph{Accuracy and Unlearning Effectiveness:} Figure~\ref{fig:accuracy} shows the performance of the proposed framework and the three baseline frameworks on the MNIST, FMNIST, and CIFAR datasets. We can observe that while FedEraser, RapidTrain, and our proposed framework an achieve the similar levels of accuracy and loss rates, they perform less favorably than FedAvg. This is attributed to the calibration process associated with the unlearning requests. Additionally, the three frameworks have comparable unlearning effectiveness with FedAvg.

\begin{figure}[!t]
    \centering
    \includegraphics[width=0.5\textwidth]{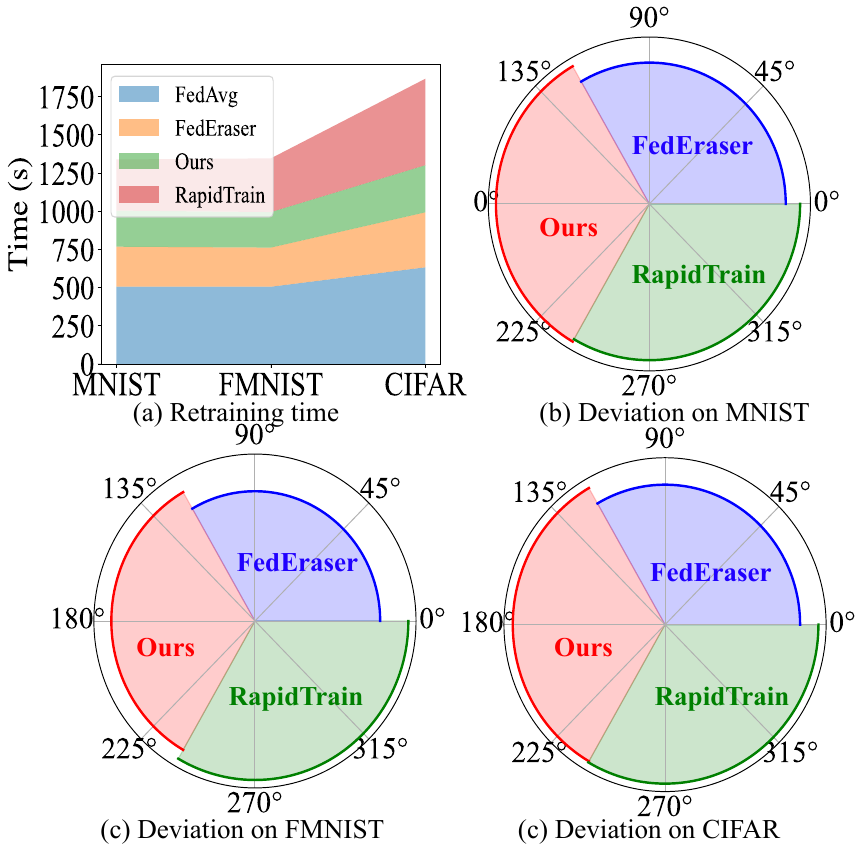}
    \caption{Time consumption and model deviation.}
    \label{fig:deviation}
\end{figure}

\paragraph{Time and Model Deviation:} We conduct experiments to validate the time analysis in Section~\ref{sec:time_analysis}. The results are shown in Figure~\ref{fig:deviation} (a). Specifically, our proposed framework achieves the shortest retraining time among all frameworks because of its refinement of adaptive retraining global rounds. Moreover, the retraining time of RapidTrain is longer than that of both Ours and FedEraser. The reason is that RapidTrain requires more computations than other frameworks for calculating loss functions. On the other hand, we compare the model deviation between the unlearning model retrained by our proposed framework and that trained by FedAvg. As shown in Figures~\ref{fig:deviation} (b), (c), and (d), RapidTrain and our proposed framework can achieve almost the same performance, and both of them outperform FedEraser. It is attributed to their specific adaptive momentum and retraining methods, thereby leading to similar results in model deviation.

\begin{figure}[!t]
    \centering
    \includegraphics[width=0.5\textwidth]{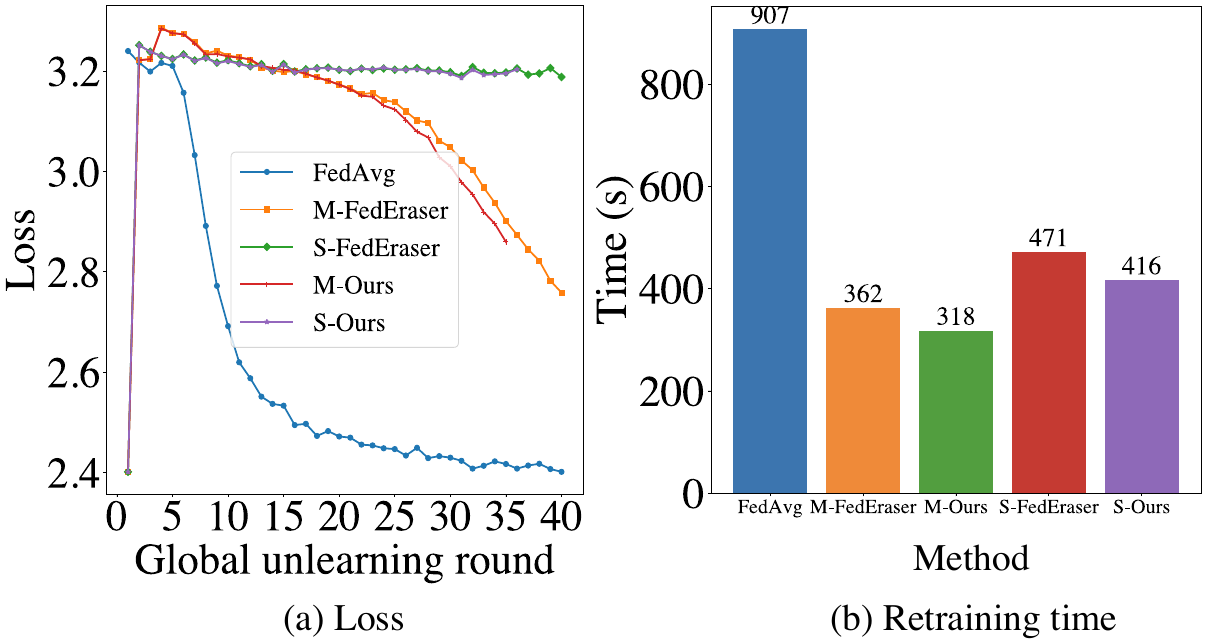}
    \caption{Performance on the generation task.}
    \label{exp:generation}
\end{figure}

\paragraph{Diversity:} To validate the diversity on the generation task with single and multiple unlearning requests, we compare the proposed framework and FedEraser on an Eminem Lyrics dataset with a NanoGPT model\footnote{https://github.com/karpathy/nanoGPT}, labeled as S-Method or M-Method. Since RapidTrain fails to converge in the generation task~\cite{su2023asynchronous}, it is not included in this experiment. We can see from Figure~\ref{exp:generation} that the proposed framework always performs better than FedEraser in terms of the retraining time without compromising the loss performance. 


\section{Conclusion}

In this paper, we propose a blockchain and Chameleon hash function-based proof of unlearning protocol to remove the data effects in FL. We also design an adaptive retraining mechanism, which evaluates the target clients' contributions to balance both model accuracy and computational overhead. Experimental results demonstrate that the shared parameters do not reside in the blockchain, and the computation overhead can be significantly reduced. For future work, considerations such as the incentive mechanism for the target clients to remove data effects, the data augmentation mechanism, and storage overhead can be explored to enhance the performance of federated unlearning. 

\bibliographystyle{named}
\bibliography{ijcai24}

\end{document}